\documentclass[letterpaper]{article} %
\usepackage[draft]{aaai25}  %
\usepackage{times}  %
\usepackage{helvet}  %
\usepackage{courier}  %
\usepackage[hyphens]{url}  %
\usepackage{graphicx} %
\def\UrlFont{\rm}  %
\usepackage{natbib}  %
\usepackage{caption} %
\usepackage{algorithm}
\usepackage{algorithmic}

\usepackage{newfloat}
\usepackage{listings}
\DeclareCaptionStyle{ruled}{labelfont=normalfont,labelsep=colon,strut=off} %
\floatstyle{ruled}
\newfloat{listing}{tb}{lst}{}
\floatname{listing}{Listing}
\usepackage{amsmath} 
\usepackage{booktabs}
\usepackage{multirow}
\usepackage{paralist}
\usepackage{xltabular}

\usepackage{amsmath,amssymb}
\usepackage{dsfont}
\usepackage{rotating}

\usepackage{enumitem}

\title{ExDDI: Explaining Drug-Drug Interaction Predictions with Natural Language}

\author {
        Zhaoyue Sun\textsuperscript{\rm 1} \quad
        Jiazheng Li\textsuperscript{\rm 2} \quad
        Gabriele Pergola\textsuperscript{\rm 1} \quad
        Yulan He\textsuperscript{\rm 1,2,3}}
\affiliations {
    \textsuperscript{1}Department of Computer Science, University of Warwick \\
    \textsuperscript{2}Department of Informatics, King's College London \quad\quad \textsuperscript{3}The Alan Turing Institute\\
    \texttt{\{Zhaoyue.Sun, Gabriele.Pergola.1\}@warwick.ac.uk} \\
    \texttt{\{Jiazheng.Li, Yulan.He\}@kcl.ac.uk}}

\usepackage{bibentry}

\begin{document}

\maketitle

\begin{abstract}
Predicting unknown drug-drug interactions (DDIs) is crucial for improving medication safety. Previous efforts in DDI prediction have typically focused on binary classification or predicting DDI categories, with the absence of explanatory insights that could enhance trust in these predictions. In this work, we propose to generate natural language explanations for DDI predictions, enabling the model to reveal the underlying pharmacodynamics and pharmacokinetics mechanisms simultaneously as making the prediction. To do this, we have collected DDI explanations from DDInter and DrugBank and developed various models for extensive experiments and analysis. Our models can provide accurate explanations for unknown DDIs between known drugs. %
This paper contributes new tools to the field of DDI prediction and lays a solid foundation for further research on generating explanations for DDI predictions.\footnote{ Code and data will be made publicly available.}
\end{abstract}

\section{Introduction}

Drug-drug interaction (DDI) refers to the alteration of the effects of one or more drugs when drugs are taken simultaneously \citep{zhang2023application}. Such changes may lead to loss of therapeutic effect or occurrence of toxicity, threatening patient safety \citep{zhang2023application}. With the increasing number of approved drugs in recent years, the likelihood of interactions between drugs has also increased \citep{khori2011frequency, han2022review}. Although wet lab experiments are available for validating DDIs, they are hindered by strict experimental conditions and high costs \citep{safdari2016computerized}, making it unfeasible to explore all potential interaction combinations. Therefore, computational methods for predicting DDIs have been extensively researched, and numerous models have demonstrated strong predictive capabilities. However, as predictive capabilities advance, models tend to become more complex and opaque, obstructing users' understanding of the predicted results \citep{vo2022road}.

\begin{figure}[ht]
    \centering
    \includegraphics[width=\linewidth]{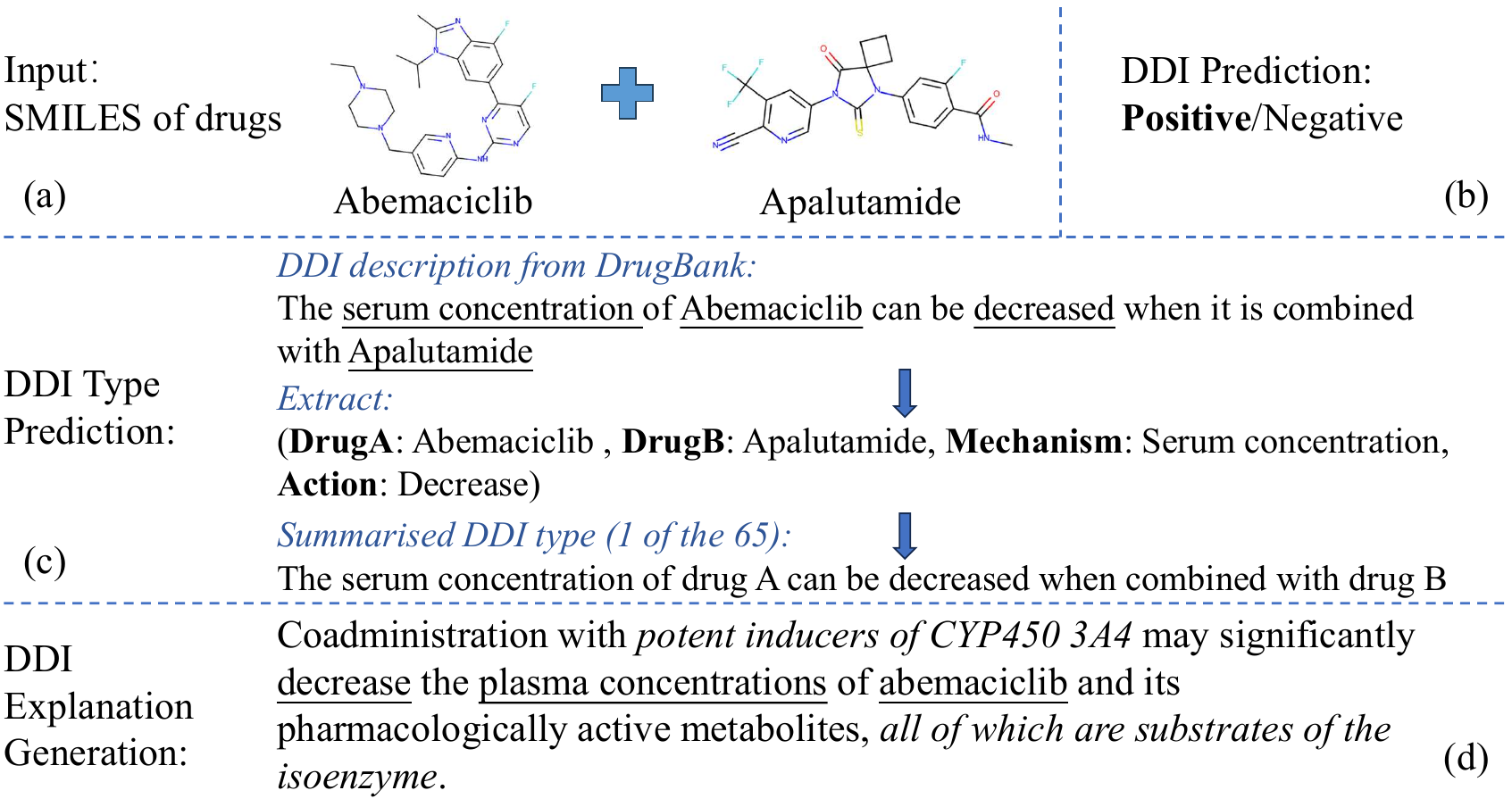}
    \caption{Examples of different DDI prediction tasks. (a) Model inputs, i.e., SMILES representations of the drug pairs; (b) Traditional DDI prediction: binary classification task; (c) DDI-type prediction: multiple classification task; (d) DDI explanation generation: our proposed task, formulated as text generation. The underlined content represents the annotations involved in DDI type prediction, while the italicized text denotes unique content provided by DDInter's explanations.}
    \label{fig:example}
\end{figure}
Specifically, the majority of previous methods have focused only on binary classification, i.e., 
predicting whether there is an interaction between two drugs (Figure 1(b)), yet overlooking the mechanisms and outcomes of DDIs \citep{zhang2023application}. To help users better grasp DDI knowledge from predictions, some studies have proposed DDI-type prediction \citep{ryu2018deep, deng2020multimodal, lin2022mdf}, which is defined as a multi-classification problem that categorises DDIs into various subtypes according to their effects. For example, \citet{deng2020multimodal} used NLP techniques to extract quadruples \texttt{(drugA, drugB, mechanism, action)} from DDI descriptions collected from DrugBank, where `\texttt{mechanism}' refers to the drugs' effects on metabolism, serum concentration, therapeutic efficacy, etc., and `\texttt{action}' indicates an increase or decrease. %
They summarised DDIs into 65 types based on the extracted quadruples used for classification (Figure \ref{fig:example}(c)).

While DDI-type prediction reveals the outcomes of DDI events, the granularity is coarse, lacking attention to the underlying causes of DDIs. As a valuable resource, \citet{xiong2022ddinter} constructed the DDInter database, gathering information on 1.8k approved drugs and 0.24M associated DDIs, along with detailed explanations. The explanations were collected from scientific literature in PubMed and medication guides of drugs, and reviewed by a clinical pharmacist team. Compared to DDI types defined by previous research, the DDI explanations provided in DDInter are more informative, encompassing not only the consequences of DDIs but also the cause mechanisms contributing to their occurrence, as illustrated in Figure \ref{fig:example}(d). 

In this work, we propose a novel task of generating natural language explanations for DDI predictions. \textbf{Our goals} are :  \begin{inparaenum}[1)]
\item to explore methods that generate explanations of the underlying pharmacodynamic or pharmacokinetic mechanisms when predicting DDIs. These explanations could help researchers evaluate the plausibility of the model’s predictions based on their expertise;
\item to investigate how the explanation generation process influences the prediction task.
\end{inparaenum}

For this task, we collected long and short explanations of DDIs from DDInter and DrugBank, respectively. We conducted extensive experiments in both \textit{transductive} and \textit{inductive} settings to meet the needs of application scenarios. We propose and evaluate the performance of the \textbf{ExDDI} family methods for DDI explanation generation, which includes \textit{three different fine-tuning paradigms}—namely, seq-to-seq, multi-task training, and multi-task training with staged inference —along with a \textit{retrieval-based unsupervised model} and an \textit{LLM-based (i.e., ChatGPT) in-context demonstration prompting model}.

\textbf{Our contributions} are: \begin{itemize}[label=•, itemsep=0pt, leftmargin=20pt]
        \item To the best of our knowledge, \textbf{\emph{we are the first to explore the DDI explanation generation task}}, which is crucial for trustworthy AI-driven drug safety research. We created the ExDDI model family for this task and carried out a comprehensive evaluation, offering tools and baselines for future studies.
        \item Our experiments reveal that \textbf{\emph{top-performed fine-tuning methods can effectively capture molecular similarities and generate accurate explanations in the transductive setting}}. However, their ability to generalise to unseen drugs during training is limited, likely due to the constraints of the linearised representation of SMILES. Fingerprint similarity-based retrieval methods can match the performance of fine-tuning approaches when both query drugs are unseen, even though they perform less effectively in settings that require less generalisation. On the other hand, general LLMs exhibit very limited capability in DDI prediction when given molecular representations of drugs. 
        \item Additionally, we demonstrate that models trained on DDInter outperform those trained on DrugBank in prediction tasks, suggesting that \textbf{\emph{rich, detailed explanations not only enhance human understanding but also improve model prediction capabilities}}. Our experimental analysis provides valuable insights for advancing future DDI-related research.
\end{itemize}

\begin{figure*}[thp] 
\centering
\includegraphics[width=1.47\columnwidth]{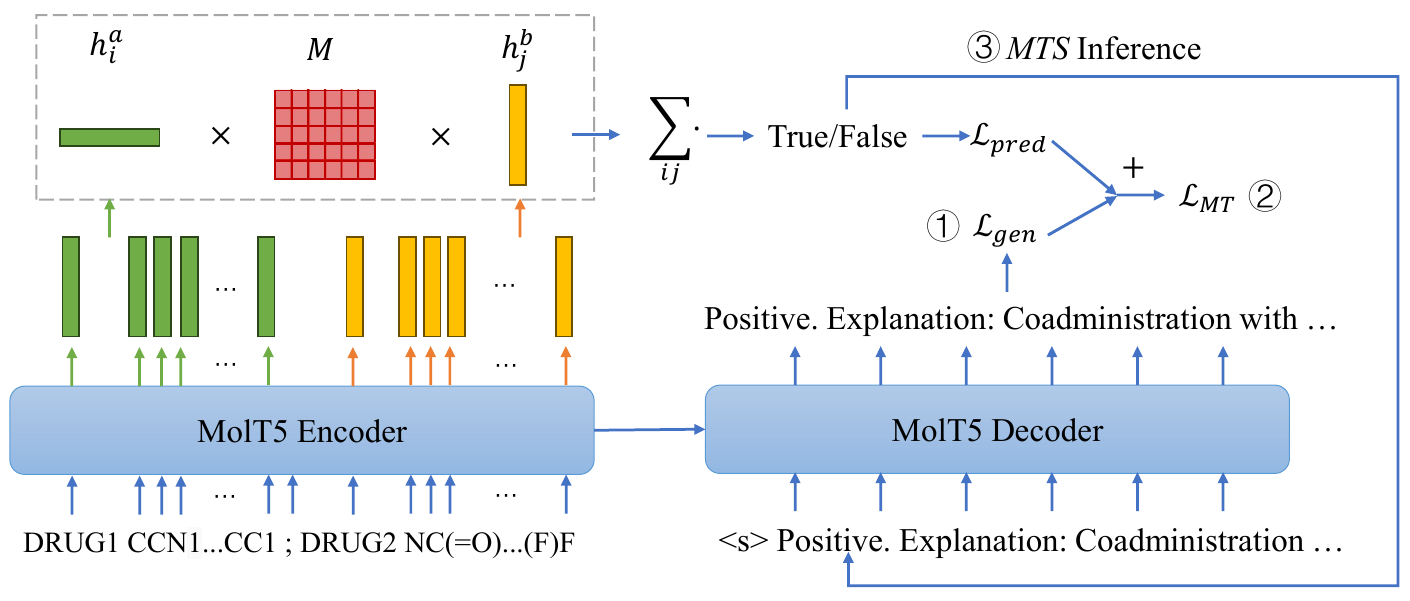} 
\caption{Illustration of the fine-tuning methods. \textcircled{1} the learning objective of the ExDDI-S2S model; \textcircled{2} the learning objective of the ExDDI-MT model; \textcircled{3} the inference step of the ExDDI-MTS model.} 
\label{fig:finetune-arch} 
\end{figure*}

\section{Related Work}

\paragraph{DDI Prediction and Interpretability}

Many efforts have been dedicated to DDI prediction over the years. Some of them are based on similarity measurements, which are grounded on the assumption that similar drugs may possess similar biological activity. Various similarity matrices - targeting molecule structure, side effect, protein targets, etc. - can be used for direct matching \citep{vilar2012drug, ferdousi2017computational} or as features to train machine learning classifiers \citep{gottlieb2012indi, cheng2014machine, sridhar2016probabilistic} and neural networks \citep{rohani2019drug, lee2019novel, zhang2022cnn}. 
Other approaches involve matrix decomposition of known DDI matrices combined with multiple relation matrices to predict unknown DDIs \citep{zhang2018manifold, rohani2020iscmf}. Additionally, recent advancements have incorporated knowledge graphs \citep{asada2023integrating, ren2022biodkg} and graph neural networks for learning single or paired molecular structures \citep{baitai2023molecular, li2023dsn, nyamabo2022drug} to enhance prediction accuracy. 

In recent years, the transparency of DDI prediction models has gained significant attention. Some studies have employed matrix factorization \citep{zhu2022multi} or attention mechanisms \citep{ma2023dual, li2023dsn} to identify representative features or substructures in DDI interactions, offering valuable insights into the underlying prediction mechanisms. However, generating natural language explanations that focus on elucidating the pharmacological principles of DDIs offers another promising direction for further exploration. Additionally, exploring whether introducing supervision signals from these explanations could enhance the prediction task itself is an intriguing question. Furthermore, natural language explanations are more user-friendly for human understanding and could be integrated with substructure-highlighting methods in future work.

\paragraph{Natural Language Explanation Generation}

Natural language explanation generation aims to create free-text explanations for model predictions to help users better understand model behaviour and make decisions. Previous work had explored various training paradigms over prediction and explanation generation, which were categorised into four types by \citet{hase-etal-2020-leakage} based on whether the model is provided with labels (\textbf{RA}) or not (\textbf{RE}) during the generation of explanations and whether the generated explanations are used as part of the input for predicting (\textbf{ST}) or not (\textbf{MT}). Specifically, the \textbf{ST-RE} paradigm in the first stage trains the model to generate explanations based on the input text and, in the second stage, learns to predict labels based on explanations generated in the first stage \citep{rajani-etal-2019-explain}. The \textbf{ST-RA} paradigm first learns to produce explanations based on labels and input, then generates an explanation for each label in the second stage and trains the model to make predictions based on all explanations \citep{hase-etal-2020-leakage}. The \textbf{MT-RE} paradigm refers to jointly training the model to generate both labels and explanations simultaneously \citep{narang2020wt5, yordanov-etal-2022-shot}. The \textbf{MT-RA} paradigm not only jointly trains the prediction and explanation generation, but also provides labels during the explanation generation process \citep{camburu2018snli, li-etal-2023-distilling}. This is achieved by feeding the gold label to train the explanation generation model and using the label predicted by the model for inference. For the DDI explanation generation task, explanations for negative cases are naturally absent. We use artificially constructed explanations for negative cases to train the model, but the relationship between such explanations and predictions is almost lexically distinguishable. Therefore, the ST paradigm is relatively less meaningful for our task. Our finetuning methods explore variants of MT-RE and MT-RA paradigms.

In recent years, inspired by the astonishing reasoning capabilities of LLMs, people have also explored generating explanations by prompting LLMs with different strategies \citep{wei2022chain, lampinen-etal-2022-language, wiegreffe-etal-2022-reframing}. However, in our interactions with general LLMs, such as ChatGPT \citep{chatgpt}, we find that while they can make reasonably sound judgments and explanations about whether known drugs have a DDI \citep{al2023evaluating}, they often express incapability when asked with chemical molecular structures (e.g., with SMILES representation). Therefore, to compare the LLM performance with other methods proposed in this work, we prompt LLMs with in-context demonstrations of several similar drug pairs.

\section{Method}

\paragraph{Task Formulation} Given a drug pair $(d_1, d_2)$, one of our objectives is to predict DDI label $l \in \{``positive", ``negative"\}$, denoting the presence or absence of interactions between these drugs when administered together. Additionally, we aim to generate a textual explanation, $s$, elucidating the rationale behind the existence or non-existence of DDIs. For positive instances, we rely on DDI descriptions sourced from DDInter \citep{xiong2022ddinter} or DrugBank as the target explanation, while for negative instances lacking natural language explanations, we formulate target explanations using a predefined template: $s$ = `\texttt{<DRUG1\_DEF>. <DRUG2\_DEF>. There were no known direct interactions reported between them.}', where \texttt{<DRUG1\_DEF>} and \texttt{<DRUG2\_DEF>} represent the drug descriptions retrieved from the DDInter database for $d_1$ and $d_2$ respectively.

We explored three setups: \emph{fine-tuning methods} with different paradigms, \emph{retrieval-based} methods, and \emph{LLM-based in-context demonstration prompting} methods on this task.

\subsection{Fine-tuning Methods}\label{sec:finetune}

For the fine-tuning methods, we constructed a Seq-to-Seq model (\textbf{ExDDI-S2S}), a Multi-Task training model with an additional classifier (\textbf{ExDDI-MT}) and a Multi-Task training model with Staged generation constrained by the classifier's prediction (\textbf{ExDDI-MTS}). We use MolT5 \citep{edwards-etal-2022-translation} as the backbone encoder-decoder for these models as it has been pre-trained on molecule-text translation tasks that establish a connection that maps molecular features and natural language representations into a shared space, thereby enhancing the model's generalisability. Figure \ref{fig:finetune-arch} shows the overall structure of the fine-tuning methods.

\paragraph{ExDDI-S2S} For each query drug pair $(d_1, d_2)$, we construct the model's input $\mathbf{x}$ as `\texttt{DRUG1 <SMILES1>; DRUG2 <SMILES2>}', where \texttt{<SMILES1>} and \texttt{<SMILES2>} correspond to the SMILES representations of $d_1$ and $d_2$, respectively. For the target output, we first replace mentions of drug names in the target explanations with `\texttt{DRUG1}' and `\texttt{DRUG2}' through regularised expression matching, and then construct the generation target sequence $\textbf{y}$ = `\texttt{<s> <LABEL> Explanation: <EXP> </s>}', where \texttt{<LABEL>} represents `positive' or `negative' and \texttt{<EXP>} is the preprocessed explanation text. Then the model is trained by the following text generation loss:
\begin{equation}
    \mathcal{L}_{gen} = - \sum_{i=1}^N \sum_{t=1}^T \log p(y_t^{(i)}|\mathbf{x}^{(i)}, \mathbf{y}_{<t}^{(i)}, \Theta),
\end{equation}
where $N$ represents the size of the training set, $T$ denotes the length of the target sequence, and $\Theta$ signifies the parameters of the encoder and decoder.

\paragraph{ExDDI-MT} Simultaneously generating prediction labels during the target sequence generation process could potentially divert the model's attention from learning the classification task effectively. Hence, we attempt to introduce an extra classification module for multi-task training. The design of the classification module is inspired by \citet{nyamabo2022drug}, where a linear transformation matrix $\mathbf{M}$ is learned to map the representations corresponding to $d_1$ and $d_2$ pairwise to a real-valued score, and the scores obtained for all pairs of representations are summed to make the prediction.

Specifically, suppose the encoder representations for the input corresponding to $d_1$ and $d_2$ are denoted as $\mathbf{H}_a = \{\mathbf{h}_1^a, \mathbf{h}_2^a, ..., \mathbf{h}_i^a...\}$ and $\mathbf{H}_b = \{\mathbf{h}_1^b, \mathbf{h}_2^b, ..., \mathbf{h}_j^b...\}$, respectively. %
Our objective is to learn the weights of $\mathbf{M}$ to predict the DDI label, with $\mathbf{M}$ having a dimension of $(768 \times 768)$ in our implementation:
\begin{equation}
    \hat{l} = \text{Sigmoid}(\sum_{ij}\mathbf{h}_i^a \mathbf{M} (\mathbf{h}_j^b)^\top).
\end{equation}
The classifier is then optimised by the binary cross-entropy loss, which is defined as:
\begin{equation}
    \mathcal{L}_{pred} = -\frac{1}{N} \sum_{i=1}^{N} [l_i \log(\hat{l}_i) + (1 - l_i) \log(1 - \hat{l}_i)].
\end{equation}

\noindent The ExDDI-MT paradigm then jointly optimises the generation loss and classifier prediction loss. Thus, the overall loss function is:
\begin{equation}
    \mathcal{L}_{MT} = \mathcal{L}_{gen} + \mathcal{L}_{pred}
    \label{eq:loss}
\end{equation}

\paragraph{ExDDI-MTS} Inspired by \citet{camburu2018snli}, we are interested in exploring whether using the predictions of the multi-task trained classifier, which may have better classification performance than the decoder, as an additional constraint for decoding can improve the quality of explanation generation. 

As shown in Figure \ref{fig:finetune-arch}, during the inference stage, we first utilise the fine-tuned encoder and classifier weights $M$ from ExDDI-MT to predict $\hat{l}$. If $\hat{l}$ is 1, then the decoding prefix $Pr$ is set to ``\texttt{<s> positive}"; otherwise, it is set to ``\texttt{<s> negative}". The prefix is prepended during decoding before generating the explanation, specifically: 
\begin{equation}
    \mathbf{y} = \text{Decoder}(Pr; \mathbf{x}; \Theta)
\end{equation}

\subsection{Retrieval-based Method}
\label{sec:ret-method}

For the retrieval-based method (\textbf{ExDDI-RV}), we retrieve the most similar drug pair in the training set to the query drugs, then use the DDI label and explanation of the retrieved case as the response to the query. This is based on the assumption that similar drugs often share similar pharmacological properties, and indeed, in the data we have collected, a large number of DDIs share the same explanation.

We retrieve the nearest drug pairs based on the similarity of the drugs' chemical molecular structures. 
Initially, we retrieve the top-$K$ ($K=50$) most similar drugs  for each query drug from the training set's drug list. The similarity score between two drugs is calculated by the Tanimoto coefficient (also known as Jaccard similarity) of their fingerprints, which are binary vectors indicating the presence or absence of specific chemical substructures. 
We employ RDKit to extract MACCS keys \citep{durant2002reoptimization} as the fingerprints. 
Subsequently, we pair the top-$K$ nearest neighbours of the two drugs, resulting in $K^2$ candidate drug pairs. Each candidate drug pair's similarity score is computed as the product of the similarity scores for each retrieved drug and its corresponding query drug. Following this, we filter out drug pairs that do not exist in the training set and re-rank the remaining pairs. Ultimately, the top drug pair is obtained as the retrieval result, and its label and explanation are used as the response. 

\subsection{LLM-based In-Context Prompting} \label{sec:in-context}

To assess the capabilities of general LLMs in predicting and generating explanations for DDIs based on molecular representations, we constructed an in-context demonstration prompting (\textbf{ExDDI-IC}) method. Based on the retrieval process described in the previous subsection, we retrieve the five most similar drug pairs for each test case. We then use their input, i.e., the SMILES representations, and the DDI label and explanation, as demonstrations to prompt ChatGPT to generate responses. The instructions used to prompt ChatGPT is shown in Appendix A.

\begin{table*}[htp]
\centering
\resizebox{\textwidth}{!}{%
\Large
\begin{tabular}{@{}lcccccccccccc@{}}
\toprule
 & \multicolumn{12}{c}{DrugBank} \\ \cmidrule(l){2-13} 
 & \multicolumn{4}{c}{Transductive} & \multicolumn{4}{c}{Inductive Test S2} & \multicolumn{4}{c}{Inductive Test S1} \\ \cmidrule(l){2-13} 
 & BLEU & ROUGE-1 & ROUGE-2 & ROUGE-L & BLEU & ROUGE-1 & ROUGE-2 & ROUGE-L & BLEU & ROUGE-1 & ROUGE-2 & ROUGE-L \\ \midrule
ExDDI-IC & 0.1187 & 0.2870 & 0.1192 & \multicolumn{1}{c|}{0.2257} & 0.1174 & 0.2891 & 0.1161 & \multicolumn{1}{c|}{0.2268} & 0.0674 & 0.2678 & 0.0808 & 0.2006 \\
ExDDI-RV & 0.5037 & 0.6780 & 0.5492 & \multicolumn{1}{c|}{0.6250} & 0.4555 & 0.6261 & 0.4866 & \multicolumn{1}{c|}{0.5717} & 0.2069 & 0.4557 & 0.2708 & 0.3906 \\
ExDDI-S2S & 0.9352 & 0.9410 & \textbf{0.9109} & \multicolumn{1}{c|}{\textbf{0.9321}} & 0.5209 & 0.6470 & 0.5321 & \multicolumn{1}{c|}{0.6179} & \textbf{0.2197} & 0.4451 & 0.2704 & 0.3915 \\
ExDDI-MT & \textbf{0.9447} & 0.9419 & 0.9076 & \multicolumn{1}{c|}{0.9319} & 0.5157 & 0.6519 & 0.5344 & \multicolumn{1}{c|}{0.6218} & 0.2071 & 0.4448 & 0.2701 & 0.3903 \\
ExDDI-MTS & 0.9441 & \textbf{0.9421} & 0.9073 & \multicolumn{1}{c|}{0.9319} & \textbf{0.5301} & \textbf{0.6590} & \textbf{0.5390} & \multicolumn{1}{c|}{\textbf{0.6281}} & 0.2145 & \textbf{0.4578} & \textbf{0.2791} & \textbf{0.4023} \\ \midrule
 & \multicolumn{12}{c}{DDInter} \\ \cmidrule(l){2-13} 
 & \multicolumn{4}{c}{Transductive} & \multicolumn{4}{c}{Inductive Test S2} & \multicolumn{4}{c}{Inductive Test S1} \\ \cmidrule(l){2-13} 
 & BLEU & ROUGE-1 & ROUGE-2 & ROUGE-L & BLEU & ROUGE-1 & ROUGE-2 & ROUGE-L & BLEU & ROUGE-1 & ROUGE-2 & ROUGE-L \\ \midrule
ExDDI-IC & 0.1367 & 0.2773 & 0.1301 & \multicolumn{1}{c|}{0.2182} & 0.1257 & 0.2687 & 0.1184 & \multicolumn{1}{c|}{0.2097} & 0.0657 & 0.2351 & 0.0715 & 0.1719 \\
ExDDI-RV & 0.5701 & 0.6456 & 0.5612 & \multicolumn{1}{c|}{0.6080} & 0.4934 & 0.5780 & 0.4825 & \multicolumn{1}{c|}{0.5376} & \textbf{0.2500} & 0.3825 & 0.2462 & 0.3305 \\
ExDDI-S2S & \textbf{0.9392} & \textbf{0.9489} & \textbf{0.9371} & \multicolumn{1}{c|}{\textbf{0.9443}} & \textbf{0.5261} & \textbf{0.6106} & \textbf{0.5249} & \multicolumn{1}{c|}{\textbf{0.5868}} & 0.2403 & \textbf{0.3877} & \textbf{0.2481} & \textbf{0.3412} \\
ExDDI-MT & 0.9383 & 0.9477 & 0.9352 & \multicolumn{1}{c|}{0.9428} & 0.5161 & 0.6041 & 0.5160 & \multicolumn{1}{c|}{0.5791} & 0.2309 & 0.3787 & 0.2365 & 0.3304 \\
ExDDI-MTS & 0.9384 & 0.9474 & 0.9350 & \multicolumn{1}{c|}{0.9425} & 0.5172 & 0.6035 & 0.5151 & \multicolumn{1}{c|}{0.5783} & 0.2309 & 0.3778 & 0.2353 & 0.3294 \\ \bottomrule
\end{tabular}%
}

\caption{Explanation generation results. Mean values over 5-fold cross-validation are presented for all models except ExDDI-IC, which was run only once due to the high cost of API calls and its low performance. The best results for each dataset are highlighted in bold. A complete table with standard deviations is provided in Appendix D to save page space.}
\label{tab:exp-rst}
\end{table*}

\begin{figure*}[t]
    \centering
    \includegraphics[width=0.8\linewidth]{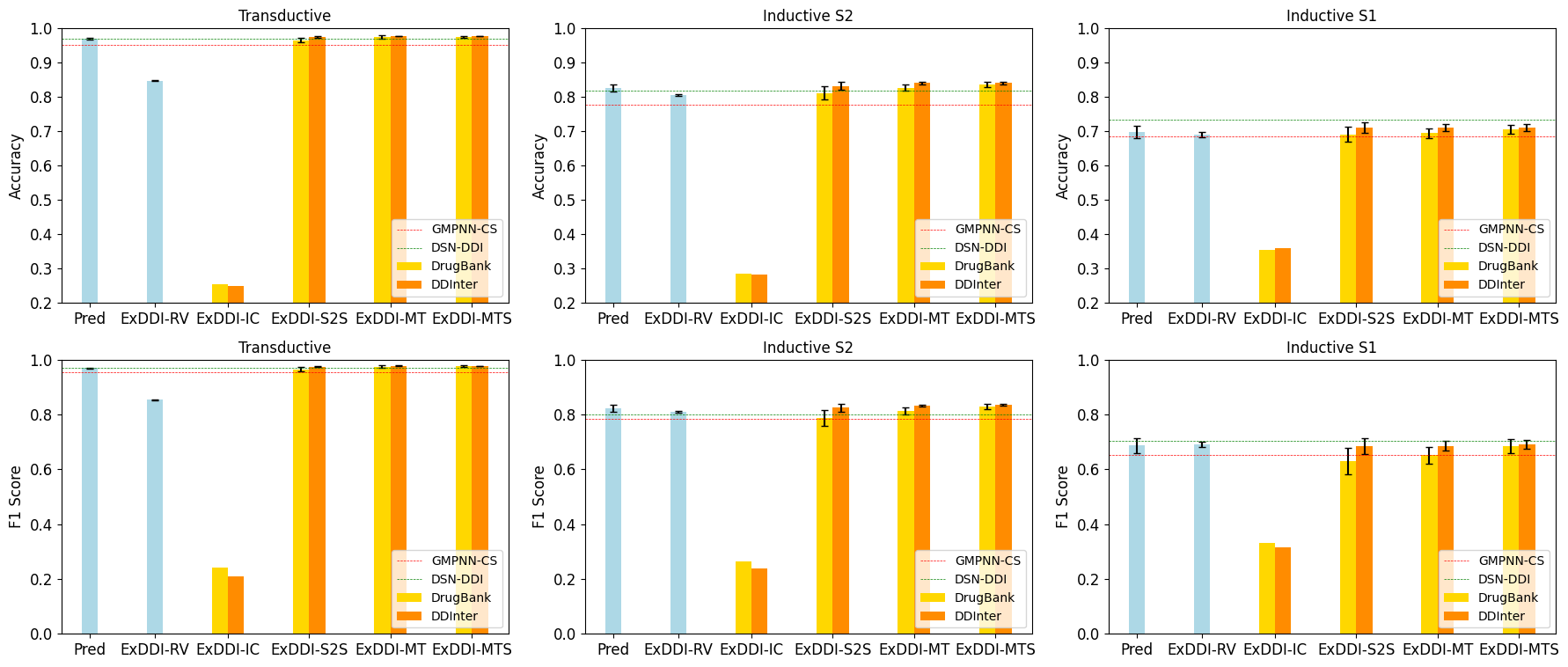}
    \caption{Binary classification results. Green and red dashed lines indicate results from the GMPNN-CS and DSN-DDI papers, respectively. Gold and orange bars represent models trained with DrugBank or DDInter explanations. Mean values over 5-fold cross-validation are shown for all models except ExDDI-IC. The error bars represent the standard deviation. Detailed numbers with precision and recall scores are available in the Appendix D.}
    \label{fig:binary-cls}
\end{figure*}

\begin{figure*}[ht]
    \centering
    \includegraphics[width=0.8\linewidth]{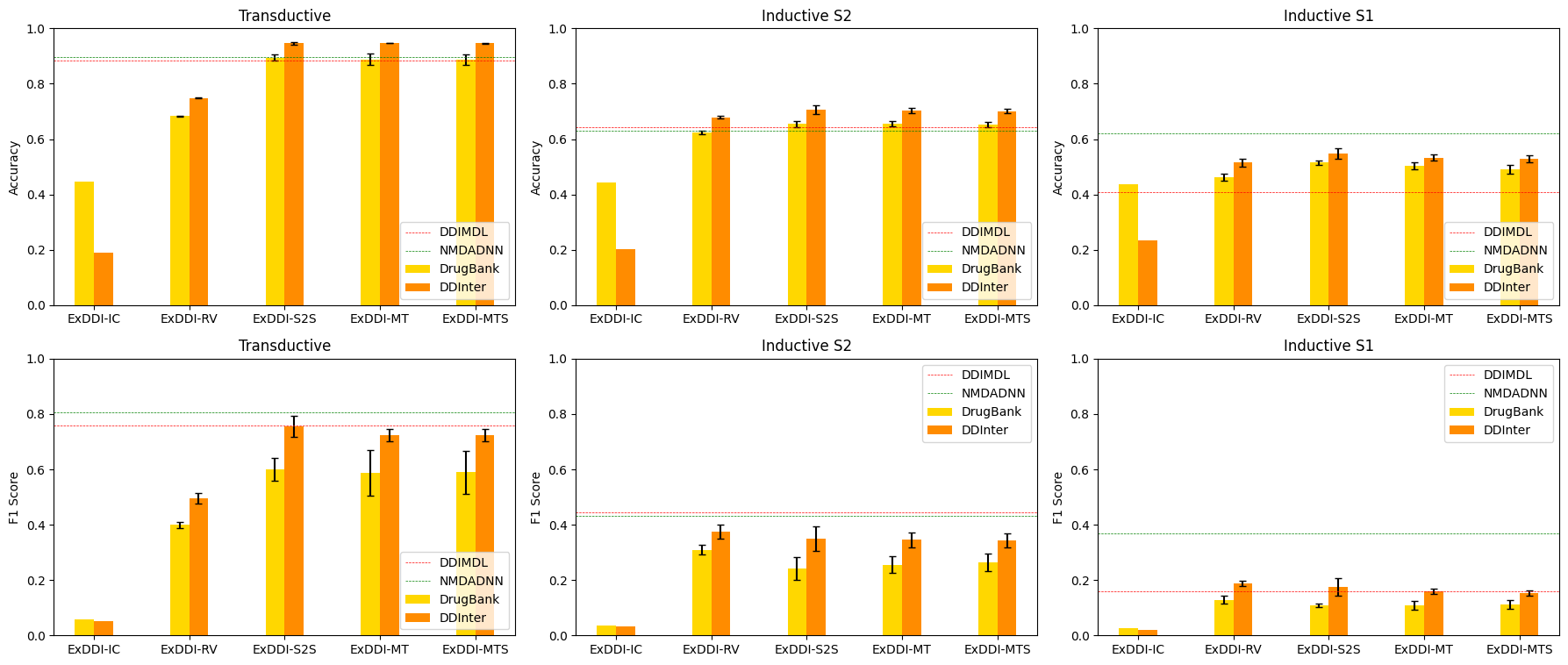}
    \caption{Multiple classification results. Green and red dashed lines indicate results from the DDIMDL and NMDADNN papers, respectively. Gold and orange bars represent models trained with DrugBank or DDInter explanations. Mean values over 5-fold cross-validation are shown for all models except ExDDI-IC. The error bars represent the standard deviation. Detailed numbers with precision and recall scores are available in the Appendix D.}
    \label{fig:mult-cls}
\end{figure*}

\section{Experiments}

\subsection{Experimental Setup}

For hyper-parameter selection and training details, please refer to Appendix B.

\paragraph{Datasets}

We evaluate the model performance based on two databases: DDInter \citep{xiong2022ddinter} and DrugBank (v5.1.10). DrugBank is a widely used resource for training and evaluating DDI prediction models, but it only provides brief DDI explanations for open download. On the other hand, DDInter offers more extensive and detailed explanations involving pharmacodynamics and pharmacokinetics principles. The selection of these two datasets is motivated by their coverage of explanations with varying lengths, enabling us to gain insights into the model's performance when generating explanations of different complexities. We collected data on drug SMILES representations, drug descriptions, annotations of DDIs and relevant explanations to construct the datasets. The detailed description and statistics of the data are reported in Appendix C.

\paragraph{Settings for Model Generalisation Evaluation}
To examine the model's generalisation ability to new drugs, we followed previous work \citep{nyamabo2022drug, li2023dsn} to evaluate the model under both transductive and inductive settings. For \textbf{\emph{transductive setting}}, we evaluate the models' performance on unknown DDI pairs, allowing drugs from the training set to also appear in the test set. We randomly divided all positive and negative samples into training/validation/test sets with a ratio of 0.7/0.1/0.2. For \textbf{\emph{inductive setting}}, we evaluate the model's performance not only on unknown DDIs but also on unknown drugs. Specifically, the test set is split into \emph{inductive S1} and \emph{inductive S2} subsets according to whether both drugs are unavailable in the training set or only one drug is unavailable in the training set. We first divided drugs into three sets, M1, M2, and M3, with proportions of 0.75/0.05/0.2. Then, the training set consists of DDI samples where both drugs in the queried drug pair are from M1; The validation set includes samples where both drugs are from M2, or one is from M2 and the other is from M1; 
The \emph{inductive S1} test set contains samples where both drugs in the drug pair are from M3, and the \emph{inductive S2} test set contains samples where one drug is from M1 and the other is from M3. We conduct 5-fold cross-validation for all settings.

\paragraph{Settings for Prediction Evaluation}

Although our primary objective of this work is to study methods for generating explanations for DDI prediction, we are also interested in exploring how the generation of explanations impacts the DDI prediction tasks.

Therefore, for \textbf{\emph{binary prediction}}, we introduce an ablation study to investigate the model's prediction performance without generating explanations, where we remove the generation loss in Eq. \ref{eq:loss} and train the model for binary classification prediction only. We also use the results reported by \citet{nyamabo2022drug} (GMPNN-CS) and \citet{li2023dsn} (DSN-DDI) as additional baselines for comparison.

For \textbf{\emph{DDI-type prediction}}, since our model does not directly learn from multi-class tasks, we estimate its performance by mapping the generated explanations to mechanism categories. Specifically, for the DrugBank data, we processed the model-generated explanations using scripts from \citet{xiong2022ddinter}'s work and mapped the extracted quadruples to known categories. For the DDInter data, we first mapped the model-generated explanations to preprocessed DDInter explanation templates based on the nearest Levenshtein distance, and then corresponded the generated explanations to DrugBank explanation categories based on the statistical relationship between DDInter explanation templates and DrugBank explanation categories. This aims to enable comparison of the results across both datasets. We report DDI-type prediction results reported by \citet{yan2021prediction} (NMDADNN) and \citet{xiong2022ddinter} (DDIMDL) for reference. However, it's worth noting that although the data sources are all DrugBank, there are differences between our constructed data and theirs, which may lead to discrepancies in explanation categories. As a result, the DDI type categories covered by the data used by \citet{yan2021prediction} and \citet{xiong2022ddinter} are 65, while the number of DDI type categories extracted using the same method on our data is 257. Additionally, NMDADNN and DDIMDL utilised additional information, such as drug targets and enzymes, in their inputs. However, considering the context of predicting DDIs for new drugs, where such knowledge might be unknown, our method utilises the chemical molecular representations of drugs as inputs only. In light of these factors, our model faces greater challenges in multi-classification tasks, which should be considered when comparing models.

\paragraph{Evaluation Metrics}

We evaluate DDI explanation generation using metrics including BLEU, ROUGE-1, ROUGE-2, and ROUGE-L scores. To assess the models' classification performance, we present accuracy and F1 scores in Figure \ref{fig:binary-cls} and Figure \ref{fig:mult-cls}, while full numerical values, including precision and recall scores, are provided in the Appendix D.  For multi-class classification, we employ macro-averaging for precision, recall, and F1 scores.

\subsection{Main Result: Explanation Generation} Table \ref{tab:exp-rst} presents the evaluation results of the models in explanation generation. The results show that \begin{inparaenum}[1)]
\item For all methods except ExDDI-IC, performance in the \textit{transductive setting} is significantly better than in the \textit{inductive S2 setting}, which in turn outperforms the \textit{inductive S1 setting}. This highlights that existing models face substantial challenges when dealing with unseen molecules. ExDDI-IC performs poorly in all three settings, indicating that general LLMs still struggle to handle molecular representations effectively.
\item In the \textit{transductive setting}, fine-tuning methods demonstrate a clear advantage over the other two categories, achieving promising results with scores exceeding 0.9. However, in the \textit{inductive S2 setting}, the advantage of fine-tuning methods over retrieval methods is greatly reduced. In the \textit{inductive S1 setting}, their performance is nearly equivalent to that of retrieval methods. This suggests that fine-tuning methods can effectively learn the similarity between drug molecules, but their generalisation ability to unseen molecules is still relatively poor. 
\item There are no statistically significant differences in scores across the three fine-tuning paradigms. Even the simplest seq-to-seq scheme can produce competitive results. However, the standard deviation of results from multi-task training is relatively smaller (as reported in Appendix D), suggesting that they may offer greater stability.
\item Although the explanations in the DDInter data are richer and longer than those in DrugBank, there is no significant difference between their evaluation scores, indicating that text length is not the primary challenge in generating DDI explanations.
\end{inparaenum}

\subsection{Prediction Performance of ExDDI Models}
\paragraph{Binary Classification} Figure \ref{fig:binary-cls} presents the performance of different models in the DDI binary prediction task. The results suggest that: \begin{inparaenum}[1)]
\item For all fine-tuning methods, the prediction results of models trained on DDInter data outperform those trained on DrugBank across all settings ($p < 0.05$, paired t-test), and the standard deviations of the scores are also smaller. This indicates that training models to generate more detailed DDI explanations can indeed be beneficial for the prediction task. As for ExDDI-IC, in most cases, the prediction performance is actually better when DrugBank explanations are provided as demonstrations compared to DDInter demonstrations. This may reveal the limited reasoning ability of general LLMs when dealing with complex explanations in the DDI task.
\item Compared to the \textbf{Pred} model trained without generation loss, i.e., by removing the decoder in Figure \ref{fig:finetune-arch}, the ExDDI-MT and ExDDI-MTS models trained on DDInter data generally perform better. However, the models trained on DrugBank data sometimes perform worse than the \textbf{Pred} model, indicating that overly brief explanations may not contribute to the prediction task. The ExDDI-S2S model, when trained on DrugBank data, may suffer more performance loss and exhibit greater fluctuations, suggesting that introducing a component-aware interaction module in multi-task training is beneficial.
\item Compared to previous state-of-the-art methods, our top-performing model is generally on par. Specifically, all our fine-tuning methods perform well in both the \textit{transductive setting} and the \textit{inductive S2 setting}, but they slightly underperform compared to DSN-DDI in the more demanding \textit{inductive S1 setting}. This drop in performance suggests that, compared to prediction methods based on graph neural networks that learn 2D molecular features, using linear SMILES representation exhibits relatively poorer generalisation when dealing with unknown drugs, despite the potential benefits of supervision signals from explanations. However, it is important to note that effectively mapping 2D molecular representations learned through graph structures into a shared space with language representations to generate well-generalised explanations has not been thoroughly explored yet, making it a promising research direction for future work.
\end{inparaenum}

\paragraph{Multiple Classification} Figure \ref{fig:mult-cls} shows the results of the model's DDI-type prediction. The results reveal that: \begin{inparaenum}[1)]
\item Retrieval and fine-tuning methods based on DDInter explanations continue to outperform models based on DrugBank in DDI-type prediction ($p < 0.05$, paired t-test), further demonstrating the value of more detailed explanations in enhancing prediction accuracy. In addition, the performance gap between ExDDI-IC, when using DDInter demonstrations versus DrugBank demonstrations, is more significant. This suggests that ChatGPT faces greater challenges in reasoning through complex explanations, particularly for more demanding tasks.
\item Fine-tuning methods perform better than retrieval methods in both the \textit{transductive setting} and the \textit{inductive S2 setting}. However, in the \textit{inductive S1 setting}, their performance degrades to a level similar to that of retrieval methods, with the F1 score potentially even lower. Among the fine-tuning methods, ExDDI-S2S shows a slight edge in terms of mean values, albeit with a larger standard deviation in most cases. These results are consistent with our analysis of the explanation generation outcomes and suggest a degree of interdependence between the two tasks.
\item Numerically, our method does not always achieve higher scores in the multi-classification task compared to previous methods. However, this may be due to our data encompassing a broader range of DDI-type labels when applying the same pre-processing script. Furthermore, the DDIMDL and NMDADNN methods utilise additional information beyond molecular expressions, such as target and enzyme data, which complicates direct comparisons.
\end{inparaenum}

\subsection{Qualitative Analysis} To evaluate the quality of the model-generated explanations from multiple perspectives, we manually evaluated 20 data points from one fold of each setting in the DDInter dataset. The examples and results are detailed in Appendix E. 

In summary, in the \textit{transductive setting}, the fine-tuning models not only achieve good prediction results, but also generate explanations that are largely consistent with the annotations. The retrieval model, while generally accurate in its predictions, occasionally produces explanations that diverge from the annotations but remain relevant. The in-context prompting model, however, shows poor prediction outcomes and generates less relevant explanations. 

In the \textit{inductive S2 setting}, both the fine-tuning and retrieval models exhibit an increased likelihood of not predicting positive cases. According to their generated explanation, this is attributed to the models only correctly learning the characteristics of one of the drugs involved. Even among correctly predicted positive cases, the number of explanations that do not fully match the annotations increases, although they still retain some degree of relevance. For negative case explanations, although most models predict correctly, they tend to generate descriptions that are accurate  for just one of drugs. Nevertheless, models trained with multi-tasking are more likely to generate descriptions that are both accurate and relevant for both drugs. Unlike in the \textit{transductive setting}, the ExDDI-IC model shows a slight improvement in correct predictions, and for false positive predictions, a significant proportion remains relevant to the actual drug interactions, whereas false negative cases often fail to provide any useful information. 

In the \textit{inductive S1 setting}, the proportion of irrelevant explanations generated by the models is higher. Here, when predictions are correct, explanations generated by seq-to-seq models are more relevant than those from multi-task models, which is the opposite of what is observed in the \textit{inductive S2 setting}. When predictions are incorrect, nearly all models fail to generate relevant explanations or only generate explanations related to one of the drugs.

\section{Conclusion}

This work introduces the task of generating natural language explanations for DDI predictions, advancing DDI computational methods towards a more trustworthy AI direction. We developed the ExDDI family of models for this task and conducted a thorough evaluation, providing tools and baselines for future research. The experimental results reveal that the top-performing methods can effectively predict and explain new DDI relationships for known drugs, but their ability to predict and explain DDIs involving new drugs still requires significant improvement. Additionally, our experiments indicate that training models to generate more detailed DDI explanations can enhance the prediction task itself. We believe that generalising to molecular structures unseen during training is a significant challenge for current DDI prediction and explanation generation models. Future research should consider incorporating the graph structure of chemical molecules and utilising multi-dimensional similarity information to learn more informative drug representations.

\section*{Acknowledgements}
This work was supported in part by the UK Engineering and Physical Sciences Research Council (EPSRC) through a Turing AI Fellowship (grant no. EP/V020579/1, EP/V020579/2).

\bibliography{custom}
\newpage

\appendix
\setcounter{table}{0}
\renewcommand{\thetable}{A\arabic{table}}
\setcounter{figure}{0}
\renewcommand{\thefigure}{A\arabic{figure}}

\section{Appendix A: Prompts for ExDDI-IC}
\label{sec:apdx-prompt}

We used the following instructions along with demonstrations to prompt ChatGPT to generate the DDI label and explanations for a query drug pair: ``\textit{Analyze whether there exists a drug-drug interaction between the query molecules, and explain the reasons. Several examples have been given for reference, and you should consider the similarity of the molecular structures between the given examples and the query molecules. First, answer Yes/No, and then explain the reasons.}''.

\section{Appendix B: Implementation Details}
\label{sec:apdx-impl}
For the fine-tuning models, we employed the MolT5-base \citep{edwards-etal-2022-translation} model, which has 220 million parameters, as the backbone. For the in-context prompting method, we utilised GPT-3.5 (turbo-0125).

For the training of fine-tuning models, we set the batch size to 64. We empirically selected the learning rate from options \{5e-3, 5e-4, 5e-5\} and consistently applied a rate of 5e-4 across all experiments. We employed a warmup ratio of 0.06 with a linear learning rate scheduler. The models were trained for 50 epochs, and training was early-stopped if the evaluation metric showed no improvement for 3 consecutive epochs. For explanation generation, we leveraged beam search with a beam size of 3.

We obey the licenses of all base models used. The training and evaluation of each model were conducted on a separate NVIDIA A100 80G GPU and we used bfloat16 to accelerate the training process.

\section{Appendix C: Dataset Details}

We collected the information on drugs, including their SMILES representations, drug descriptions, annotations of DDIs and relevant explanations, from two databases: DDInter \citep{xiong2022ddinter} and DrugBank (v5.1.10). DDInter contains a total of 1.8k FDA-approved drugs and 0.24M associated DDIs. Each DDI is annotated with a description that explains the consequences of DDI effects and their pharmacokinetic or pharmacodynamic mechanisms, which are referred to as DDI explanations in our work. Due to similarities between DDIs, DDI explanations may overlap; however, the dataset still covers 5,560 distinct DDI mechanisms reported by \citet{xiong2022ddinter}. DrugBank includes 4.4k drug entries and 2.8M related DDIs. However, it only provides a short description of the DDI mechanism for open download, which mainly describes the effect. We used the script provided by \citet{deng2020multimodal} to extract mechanisms from DrugBank DDI descriptions, resulting in a number of 257. To ensure data consistency, we took the intersection of positive and negative samples from DDInter and DrugBank and randomly sampled an equal number of negative examples to the positive ones when constructing the dataset. Table \ref{tab:dat_stat} shows the statistics information of the dataset.

\begin{table}[ht]
\centering
\resizebox{1\columnwidth}{!}{
\Large
\begin{tabular}{lcccc}
\hline
        & Instances & Drugs & \begin{tabular}[c]{@{}c@{}}DDI Mechanisms\\ (DrugBank)\end{tabular} & \begin{tabular}[c]{@{}c@{}}DDI Mechanisms\\ (DDInter)\end{tabular} \\ \hline
        & \multicolumn{4}{c}{Transductive Setting}                                                                                                                     \\ \hline
           
Training   & 157,047   & 1,363 & 239                                                                 & 4,544                                                              \\
Validation & 22,435    & 1,363 & 162                                                                 & 2,178                                                              \\
Test       & 44,870    & 1,363 & 188                                                                 & 2,890                                                              \\ \hline
           & \multicolumn{4}{c}{Inductive Setting}                                                                                                                        \\ \hline
Training   & 121,751   & 1,022 & 213                                                                 & 3,481                                                              \\
Validation & 17,624    & 1,090 & 133                                                                 & 1,634                                                              \\
Test S1    & 8,786     & 272   & 96                                                                  & 720                                                                \\
Test S2    & 65,981    & 1,295 & 197                                                                 & 3,070                                                              \\ \hline
\end{tabular}}
\caption{Statistics of data splits. The number of instances includes both positive and negative cases. All values are averaged across 5 folds.}

\label{tab:dat_stat}
\end{table}

\section{Appendix D: More Experimental Results}

In this section, we present tables with more detailed experimental results. Table \ref{tab:exp-rst} presents the explanation generation results, including standard deviation values from 5-fold cross-validation. Table \ref{tab:binary-cls} and Table \ref{tab:mult-cls} show the models' accuracy, F1 score, precision, recall scores, and standard deviation values for binary and multi-class predictions, respectively.

\begin{sidewaystable*}

\centering
\resizebox{\textwidth}{!}{%
\Large
\begin{tabular}{@{}lcccccccccccc@{}}
\toprule
 & \multicolumn{12}{c}{DrugBank} \\ \cmidrule(l){2-13} 
 & \multicolumn{4}{c}{Transductive} & \multicolumn{4}{c}{Inductive Test S2} & \multicolumn{4}{c}{Inductive Test S1} \\ \cmidrule(l){2-13} 
 & BLEU & ROUGE-1 & ROUGE-2 & ROUGE-L & BLEU & ROUGE-1 & ROUGE-2 & ROUGE-L & BLEU & ROUGE-1 & ROUGE-2 & ROUGE-L \\ \midrule
ExDDI-IC & 0.1187 & 0.2870 & 0.1192 & \multicolumn{1}{c|}{0.2257} & 0.1174 & 0.2891 & 0.1161 & \multicolumn{1}{c|}{0.2268} & 0.0674 & 0.2678 & 0.0808 & 0.2006 \\
ExDDI-RV & 0.5037 $\pm.0019$ & 0.6780 $\pm.0016$ & 0.5492 $\pm.0019$ & \multicolumn{1}{c|}{0.6250 $\pm.0015$} & 0.4555 $\pm.0097$ & 0.6261 $\pm.0031$ & 0.4866 $\pm.0050$ & \multicolumn{1}{c|}{0.5717 $\pm.0035$} & 0.2069 $\pm.0074$ & 0.4557 $\pm.0069$ & 0.2708 $\pm.0081$ & 0.3906 $\pm.0069$ \\
ExDDI-S2S & 0.9352 $\pm.0120$ & 0.9410 $\pm.0074$ & \textbf{0.9109 $\pm.0097$} & \multicolumn{1}{c|}{\textbf{0.9321 $\pm.0080$}} & 0.5209 $\pm.0110$ & 0.6470 $\pm.0124$ & 0.5321 $\pm.0124$ & \multicolumn{1}{c|}{0.6179 $\pm.0122$} & \textbf{0.2197 $\pm.0033$} & 0.4451 $\pm.0138$ & 0.2704 $\pm.0106$ & 0.3915 $\pm.0125$ \\
ExDDI-MT & \textbf{0.9447 $\pm.0118$} & 0.9419 $\pm.0111$ & 0.9076 $\pm.0169$ & \multicolumn{1}{c|}{0.9319 $\pm.0129$} & 0.5157 $\pm.0122$ & 0.6519 $\pm.0061$ & 0.5344 $\pm.0067$ & \multicolumn{1}{c|}{0.6218 $\pm.0061$} & 0.2071 $\pm.0094$ & 0.4448 $\pm.0113$ & 0.2701 $\pm.0097$ & 0.3903 $\pm.0104$ \\
ExDDI-MTS & 0.9441 $\pm.0062$ & \textbf{0.9421 $\pm.0099$} & 0.9073 $\pm.0158$ & \multicolumn{1}{c|}{0.9319 $\pm.0116$} & \textbf{0.5301 $\pm.0103$} & \textbf{0.6590 $\pm.0054$} & \textbf{0.5390 $\pm.0065$} & \multicolumn{1}{c|}{\textbf{0.6281 $\pm.0056$}} & 0.2145 $\pm.0078$ & \textbf{0.4578 $\pm.0110$} & \textbf{0.2791 $\pm.0093$} & \textbf{0.4023 $\pm.0099$} \\ \midrule
 & \multicolumn{12}{c}{DDInter} \\ \cmidrule(l){2-13} 
 & \multicolumn{4}{c}{Transductive} & \multicolumn{4}{c}{Inductive Test S2} & \multicolumn{4}{c}{Inductive Test S1} \\ \cmidrule(l){2-13} 
 & BLEU & ROUGE-1 & ROUGE-2 & ROUGE-L & BLEU & ROUGE-1 & ROUGE-2 & ROUGE-L & BLEU & ROUGE-1 & ROUGE-2 & ROUGE-L \\ \midrule
ExDDI-IC & 0.1367 & 0.2773 & 0.1301 & \multicolumn{1}{c|}{0.2182} & 0.1257 & 0.2687 & 0.1184 & \multicolumn{1}{c|}{0.2097} & 0.0657 & 0.2351 & 0.0715 & 0.1719 \\
ExDDI-RV & 0.5701 $\pm.0011$ & 0.6456 $\pm.0012$ & 0.5612 $\pm.0014$ & \multicolumn{1}{c|}{0.6080 $\pm.0012$} & 0.4934 $\pm.0076$ & 0.5780 $\pm.0058$ & 0.4825 $\pm.0071$ & \multicolumn{1}{c|}{0.5376 $\pm.0063$} & \textbf{0.2500 $\pm.0119$} & 0.3825 $\pm.0105$ & 0.2462 $\pm.0118$ & 0.3305 $\pm.0111$ \\
ExDDI-S2S & \textbf{0.9392 $\pm.0076$} & \textbf{0.9489 $\pm.0045$} & \textbf{0.9371 $\pm.0054$} & \multicolumn{1}{c|}{\textbf{0.9443 $\pm.0049$}} & \textbf{0.5261 $\pm.0192$} & \textbf{0.6106 $\pm.0115$} & \textbf{0.5249 $\pm.0129$} & \multicolumn{1}{c|}{\textbf{0.5868 $\pm.0121$}} & 0.2403 $\pm.0153$ & \textbf{0.3877 $\pm.0098$} & \textbf{0.2481 $\pm.0103$} & \textbf{0.3412 $\pm.0114$} \\
ExDDI-MT & 0.9383 $\pm.0015$ & 0.9477 $\pm.0006$ & 0.9352 $\pm.0008$ & \multicolumn{1}{c|}{0.9428 $\pm.0007$} & 0.5161 $\pm.0139$ & 0.6041 $\pm.0089$ & 0.5160 $\pm.0096$ & \multicolumn{1}{c|}{0.5791 $\pm.0096$} & 0.2309 $\pm.0104$ & 0.3787 $\pm.0090$ & 0.2365 $\pm.0088$ & 0.3304 $\pm.0099$ \\
ExDDI-MTS & 0.9384 $\pm.0013$ & 0.9474 $\pm.0007$ & 0.9350 $\pm.0009$ & \multicolumn{1}{c|}{0.9425 $\pm.0008$} & 0.5172 $\pm.0137$ & 0.6035 $\pm.0089$ & 0.5151 $\pm.0096$ & \multicolumn{1}{c|}{0.5783 $\pm.0096$} & 0.2309 $\pm.0104$ & 0.3778 $\pm.0102$ & 0.2353 $\pm.0099$ & 0.3294 $\pm.0110$ \\ \bottomrule
\end{tabular}%
}

\caption{Explanation generation results. Mean values and standard deviations from 5-fold cross-validation are presented for all models except ExDDI-IC, which was run only once due to the high cost of API calls and its low performance. The best results for each dataset are highlighted in \textbf{bold}. }
\label{tab:exp-rst}

    \centering
\resizebox{\textwidth}{!}{%
\begin{tabular}{@{}lcccccccccccc@{}}
\toprule
 & \multicolumn{4}{c}{Transductive} & \multicolumn{4}{c}{Inductive Test S2} & \multicolumn{4}{c}{Inductive Test S1} \\ \midrule
 & ACC & F1 & P & R & ACC & F1 & P & R & ACC & F1 & P & R \\ \midrule
GMPNN-CS* & 0.9530 {$\pm.0005$} & 0.9539 {$\pm.0005$} & 0.9360 {$\pm.0007$} & \multicolumn{1}{c|}{0.9722{$\pm.0010$}} & 0.7772 {$\pm.0030$} & 0.7829 {$\pm.0016$} & - & \multicolumn{1}{c|}{-} & 0.6857 {$\pm.0030$} & 0.6532 {$\pm.0023$} & - & - \\
DSN-DDI* & 0.9694 {$\pm.0002$} & 0.9693 {$\pm.0002$} & - & \multicolumn{1}{c|}{-} & 0.8192 {$\pm.0120$} & 0.8018 {$\pm.0149$} & - & \multicolumn{1}{c|}{-} & 0.7342 {$\pm.0129$} & 0.7034 {$\pm.0098$} & - & - \\ \midrule
ExDDI-Pred & 0.9700 {$\pm.0019$} & 0.9701{$\pm.0020$} & 0.9666{$\pm .0030$} & \multicolumn{1}{c|}{0.9737{$\pm.0027$}} & 0.8264{$\pm.0104$} & 0.8236{$\pm.0129$} & 0.8397{$\pm.0199$} & \multicolumn{1}{c|}{0.8093{$\pm.0324$}} & 0.6988{$\pm.0179$} & 0.6874{$\pm.0275$} & 0.7202{$\pm.0233$} & 0.6597{$\pm.0512$} \\
ExDDI-RV & 0.8475{$\pm.0012$} & 0.8535{$\pm.0010$} & 0.8213{$\pm.0017$} & \multicolumn{1}{c|}{0.8884{$\pm.0022$}} & 0.8047{$\pm.0029$} & 0.8101{$\pm.0038$} & 0.7902{$\pm.0042$} & \multicolumn{1}{c|}{\textbf{0.8312{$\pm.0088$}}} & 0.6902{$\pm.0079$} & \textbf{0.6908{$\pm.0099$}} & 0.6950{$\pm.0079$} & \textbf{0.6867{$\pm.0155$}} \\
ExDDI-IC (Drugbank) & 0.2530 & 0.2426 & 0.2460 & \multicolumn{1}{c|}{0.2393} & 0.2841 & 0.2647 & 0.2720 & \multicolumn{1}{c|}{0.2577} & 0.3546 & 0.3331 & 0.3446 & 0.3223 \\
ExDDI-S2S (DrugBank) & 0.9662{$\pm.0062$} & 0.9656{$\pm.0066$} & 0.9806{$\pm.0013$} & \multicolumn{1}{c|}{0.9511{$\pm.0136$}} & 0.8115{$\pm.0188$} & 0.7875{$\pm.0285$} & \textbf{0.9017{$\pm.0048$}} & \multicolumn{1}{c|}{0.7003{$\pm.0442$}} & 0.6913{$\pm.0207$} & 0.6303{$\pm.0467$} & \textbf{0.7906{$\pm.0110$}} & 0.5273{$\pm.0620$} \\
ExDDI-MT (DrugBank) & 0.9751{$\pm.0051$} & 0.9749{$\pm.0054$} & \textbf{0.9809{$\pm.0019$}} & \multicolumn{1}{c|}{0.9690{$\pm.0118$}} & 0.8272{$\pm.0095$} & 0.8133{$\pm.0138$} & 0.8866{$\pm.0107$} & \multicolumn{1}{c|}{0.7519{$\pm.0277$}} & 0.6942{$\pm.0147$} & 0.6514{$\pm.0310$} & 0.7650{$\pm.0213$} & 0.5702{$\pm.0531$} \\
ExDDI-MTS (DrugBank) & 0.9756{$\pm.0030$} & 0.9757{$\pm.0030$} & 0.9721{$\pm.0022$} & \multicolumn{1}{c|}{0.9793{$\pm.0043$}} & 0.8366{$\pm.0070$} & 0.8305{$\pm.0098$} & 0.8648{$\pm.0130$} & \multicolumn{1}{c|}{0.7995{$\pm.0242$}} & 0.7059{$\pm.0137$} & 0.6859{$\pm.0262$} & 0.7427{$\pm.0227$} & 0.6403{$\pm.0536$} \\
ExDDI-IC (DDInter) & 0.2493 & 0.2095 & 0.2212 & \multicolumn{1}{c|}{0.1990} & 0.2827 & 0.2399 & 0.2552 & \multicolumn{1}{c|}{0.2264} & 0.3575 & 0.3140 & 0.3368 & 0.2941 \\
ExDDI-S2S (DDInter) & 0.9752{$\pm.0021$} & 0.9752{$\pm.0022$} & 0.9741{$\pm.0020$} & \multicolumn{1}{c|}{0.9763{$\pm.0056$}} & 0.8323{$\pm.0120$} & 0.8247{$\pm.0150$} & 0.8658{$\pm.0135$} & \multicolumn{1}{c|}{0.7880{$\pm.0293$}} & 0.7106{$\pm.0142$} & 0.6847{$\pm.0291$} & 0.7588{$\pm.0157$} & 0.6269{$\pm.0554$} \\
ExDDI-MT (DDInter) & \textbf{0.9783{$\pm.0006$}} & \textbf{0.9783{$\pm.0006$}} & 0.9768{$\pm.0017$} & \multicolumn{1}{c|}{0.9799{$\pm.0012$}} & 0.8404{$\pm.0039$} & 0.8334{$\pm.0038$} & 0.8741{$\pm.0088$} & \multicolumn{1}{c|}{0.7964{$\pm.0056$}} & 0.7104{$\pm.0104$} & 0.6860{$\pm.0180$} & 0.7563{$\pm.0185$} & 0.6286{$\pm.0293$} \\
ExDDI-MTS (DDInter) & 0.9778{$\pm.0006$} & 0.9779{$\pm.0006$} & 0.9743{$\pm.0011$} & \multicolumn{1}{c|}{\textbf{0.9815{$\pm.0007$}}} & \textbf{0.8413{$\pm.0036$}} & \textbf{0.8353{$\pm.0032$}} & 0.8708{$\pm.0098$} & \multicolumn{1}{c|}{0.8027{$\pm.0066$}} & \textbf{0.7115{$\pm.0101$}} & 0.6904{$\pm.0155$} & 0.7524{$\pm.0230$} & 0.6392{$\pm.0281$} \\  \bottomrule
\end{tabular}%
}
\caption{Binary classification results. * indicates results obtained from the original paper. (·) denotes the datasets used for model training. Mean values and standard deviations from 5-fold cross-validation are presented for all models except ExDDI-IC. We only \textbf{bolded} the best results from our methods due to data differences from previous work.}
\label{tab:binary-cls}

\centering
\resizebox{\textwidth}{!}{%
\begin{tabular}{@{}lcccccccccccc@{}}
\toprule
 & \multicolumn{4}{c}{Transductive} & \multicolumn{4}{c}{Inductive Test S2} & \multicolumn{4}{c}{Inductive Test S1} \\ \cmidrule(l){2-13} 
 & ACC & F1 & P & R & ACC & F1 & P & R & ACC & F1 & P & R \\ \midrule
DDIMDL* & 0.8852 & 0.7585 & 0.8471 & \multicolumn{1}{c|}{0.7182} & 0.6415 & 0.4460 & 0.5607 & \multicolumn{1}{c|}{0.4319} & 0.4075 & 0.1590 & 0.2408 & 0.1452 \\
NMDADNN* & 0.8978 & 0.8067 & 0.8748 & \multicolumn{1}{c|}{0.7811} & 0.6302 & 0.4325 & 0.5408 & \multicolumn{1}{c|}{0.3835} & 0.6200 & 0.3674 & 0.3958 & 0.4006 \\ \midrule
ExDDI-IC (DrugBank) & 0.4454 & 0.0578 & 0.1134 & \multicolumn{1}{c|}{0.0494} & 0.4426 & 0.0372 & 0.0932 & \multicolumn{1}{c|}{0.0285} & 0.4352 & 0.0255 & 0.0586 & 0.0217 \\
ExDDI-RV (DrugBank) & 0.6831 $\pm .0015$ & 0.3990 $\pm .0104$ & 0.3980 $\pm .0132$ & \multicolumn{1}{c|}{0.4290 $\pm .0099$} & 0.6227 $\pm .0064$ & 0.3097 $\pm .0181$ & 0.3026 $\pm .0224$ & \multicolumn{1}{c|}{0.3695 $\pm .0162$} & 0.4624 $\pm .0125$ & 0.1283 $\pm .0145$ & 0.1317 $\pm .0152$ & 0.1459 $\pm .0167$ \\
ExDDI-S2S (DrugBank) & 0.8938 $\pm .0114$ & 0.6008 $\pm .0415$ & 0.6387 $\pm .0357$ & \multicolumn{1}{c|}{0.5971 $\pm .0455$} & 0.6544 $\pm .0103$ & 0.2420 $\pm .0404$ & 0.3120 $\pm .0389$ & \multicolumn{1}{c|}{0.2358 $\pm .0490$} & 0.5147 $\pm .0069$ & 0.1080 $\pm .0065$ & 0.1636 $\pm .0212$ & 0.0982 $\pm .0085$ \\
ExDDI-MT (DrugBank) & 0.8882 $\pm .0193$ & 0.5864 $\pm .0825$ & 0.6164 $\pm .0676$ & \multicolumn{1}{c|}{0.5934 $\pm .0908$} & 0.6555 $\pm .0084$ & 0.2560 $\pm .0295$ & 0.3104 $\pm .0453$ & \multicolumn{1}{c|}{0.2584 $\pm .0230$} & 0.5034 $\pm .0121$ & 0.1091 $\pm .0152$ & 0.1421 $\pm .0325$ & 0.1099 $\pm .0150$ \\
ExDDI-MTS (DrugBank) & 0.8870 $\pm .0185$ & 0.5896 $\pm .0773$ & 0.6136 $\pm .0669$ & \multicolumn{1}{c|}{0.6013 $\pm .0839$} & 0.6513 $\pm .0094$ & 0.2632 $\pm .0310$ & 0.3061 $\pm .0453$ & \multicolumn{1}{c|}{0.2750 $\pm .0229$} & 0.4897 $\pm .0156$ & 0.1111 $\pm .0154$ & 0.1384 $\pm .0386$ & 0.1164 $\pm .0143$ \\
ExDDI-IC (DDInter) & 0.1891 & 0.0524 & 0.0771 & \multicolumn{1}{c|}{0.0623} & 0.2035 & 0.0323 & 0.0577 & \multicolumn{1}{c|}{0.0461} & 0.2348 & 0.0193 & 0.0292 & 0.0170 \\
ExDDI-RV (DDInter) & 0.7481 $\pm .0016$ & 0.4958 $\pm .0183$ & 0.4873 $\pm .0207$ & \multicolumn{1}{c|}{0.5337 $\pm .0146$} & 0.6792 $\pm .0059$ & \textbf{0.3763 $\pm .0252$} & 0.3657 $\pm .0288$ & \multicolumn{1}{c|}{\textbf{0.4446 $\pm .0265$}} & 0.5148 $\pm .0131$ & \textbf{0.1883 $\pm .0106$} & 0.1873 $\pm .0184$ & \textbf{0.2159 $\pm .0123$} \\
ExDDI-S2S (DDInter) & 0.9462 $\pm .0050$ & \textbf{0.7551 $\pm .0372$} & \textbf{0.7817 $\pm .0363$} & \multicolumn{1}{c|}{\textbf{0.7505 $\pm .0380$}} & \textbf{0.7059 $\pm .0161$} & 0.3486 $\pm .0446$ & \textbf{0.3942 $\pm .0616$} & \multicolumn{1}{c|}{0.3641 $\pm .0436$} & \textbf{0.5472 $\pm .0199$} & 0.1755 $\pm .0321$ & \textbf{0.2115 $\pm .0529$} & 0.1851 $\pm .0207$ \\
ExDDI-MT (DDInter) & \textbf{0.9465 $\pm .0007$} & 0.7239 $\pm .0215$ & 0.7424 $\pm .0260$ & \multicolumn{1}{c|}{0.7277 $\pm .0195$} & 0.7028 $\pm .0086$ & 0.3459 $\pm .0266$ & 0.3713 $\pm .0269$ & \multicolumn{1}{c|}{0.3732 $\pm .0335$} & 0.5327 $\pm .0104$ & 0.1591 $\pm .0108$ & 0.1857 $\pm .0193$ & 0.1689 $\pm .0103$ \\
ExDDI-MTS (DDInter) & 0.9458 $\pm .0007$ & 0.7228 $\pm .0219$ & 0.7395 $\pm .0275$ & \multicolumn{1}{c|}{0.7283 $\pm .0196$} & 0.7014 $\pm .0085$ & 0.3430 $\pm .0258$ & 0.3670 $\pm .0250$ & \multicolumn{1}{c|}{0.3705 $\pm .0332$} & 0.5293 $\pm .0129$ & 0.1531 $\pm .0080$ & 0.1798 $\pm .0138$ & 0.1613 $\pm .0113$ \\ \bottomrule
\end{tabular}%
}
\caption{Multiple classification results. F1, P and R are macro-averaged. * indicates results obtained from the original paper. (·) denotes the datasets used for model training. Mean values and standard deviations from 5-fold cross-validation are presented for all models except ExDDI-IC. We only \textbf{bolded} the best results from our methods due to data differences from previous work.}
\label{tab:mult-cls}

\end{sidewaystable*}

\section{Appendix E: Human Evaluation Details}
\label{sec:apdx-human-eval}

We sampled 20 cases each from the transductive, inductive S1, and inductive S2 test sets of the DDInter data and manually examined the explanations generated by the five models. 

Specifically, for true positive (TP) and false positive (FP) predictions, we categorised the generated explanations into the following types:
\begin{itemize}
    \item $+$: The generated explanation semantically aligns with the gold standard explanation (applicable only to TPs).
    \item $\sim$: The generated explanation, while not an exact match to the gold standard, remains relevant. For FPs, this also includes the case that the explanation highlights specific properties of the drugs.
    \item $-$: The generated explanation is irrelevant to the gold standard explanation and the drugs.
\end{itemize}

For true negative (TN) and false negative (FN) predictions, where the models are expected to generate descriptions for both drugs, we assess the relevance of generated descriptions for each drug. The categories are as follows:
\begin{itemize}
    \item $++$: The generated descriptions for both drugs are semantically consistent with the gold standards.
    \item $+\sim$: The generated description for one drug semantically aligns with the gold standard drug description, while the description for the other drug exhibits only partial relevance to the gold standard. 
    \item $+-$:  The generated description for one drug is semantically consistent with the gold standard drug description, while the description for the other drug is irrelevant to the gold standard.
    \item $\sim\sim$:  The generated descriptions for both drugs are partially relevant to the gold standards.
    \item $\sim-$:  One of the generated drug descriptions is partially relevant to the gold standard, while the other is completely irrelevant.
    \item $--$: The generated descriptions for both drugs are irrelevant to the gold standard descriptions.
\end{itemize}

Tables \ref{tab:human-trans}-\ref{tab:human-s1} present the results for each dataset and Table \ref{tab:case-example} provides an example of annotation for each category.

\section{Appendix F: Ethic Statement}
Our development of the DDI explanation generation model aims to aid researchers in identifying new DDIs. However, practitioners should be aware that the model may sometimes produce erroneous outputs and should use their professional expertise to assess the reliability of the model-generated content. It is important to note that the models and data presented in this work are not intended for use as medical advice.

\begin{table*}[htb]
\resizebox{1.97\columnwidth}{!}{%
\begin{tabular}{@{}lccccccccccccccccccccc@{}}
\toprule
 & \multicolumn{4}{c}{TP} & \multicolumn{3}{c}{FP} & \multicolumn{7}{c}{TN} & \multicolumn{7}{c}{FN} \\ 
 \cmidrule(r){2-5}  \cmidrule(l){6-8} \cmidrule(l){9-15} \cmidrule(l){16-22} 
 & $+$ & $\sim$ & $-$ & Total & $\sim$ & $-$ & Total & $++$ & $+\sim$ & $+-$ & $\sim$$\sim$ & $\sim-$ & $--$ & Total & $++$ & $+\sim$ & $+-$ & $\sim$$\sim$ & $\sim-$ & $--$ & Total \\ \midrule
ExDDI-IC & 1 & 0 & 0 & \textbf{1} & 2 & 5 & \textbf{7} & 1 & 0 & 0 & 0 & 0 & 0 & \textbf{1} & 0 & 0 & 1 & 0 & 1 & 9 & \textbf{11} \\
ExDDI-RV & 6 & 4 & 1 & \textbf{11} & 0 & 1 & \textbf{1} & 2 & 1 & 3 & 0 & 0 & 1 & \textbf{7} & 0 & 1 & 0 & 0 & 0 & 0 & \textbf{1} \\
ExDDI-S2S & 11 & 0 & 1 & \textbf{12} & 0 & 0 & \textbf{0} & 8 & 0 & 0 & 0 & 0 & 0 & \textbf{8} & 0 & 0 & 0 & 0 & 0 & 0 & \textbf{0} \\
ExDDI-MT & 11 & 0 & 1 & \textbf{12} & 0 & 0 & \textbf{0} & 8 & 0 & 0 & 0 & 0 & 0 & \textbf{8} & 0 & 0 & 0 & 0 & 0 & 0 & \textbf{0} \\
ExDDI-MTS & 11 & 0 & 1 & \textbf{12} & 0 & 0 & \textbf{0} & 8 & 0 & 0 & 0 & 0 & 0 & \textbf{8} & 0 & 0 & 0 & 0 & 0 & 0 & \textbf{0} \\ \bottomrule
\end{tabular}%
}
\caption{Results of human evaluation on the DDInter transductive test set.}
\label{tab:human-trans}
\end{table*}

\begin{table*}[htb]
\resizebox{1.97\columnwidth}{!}{%
\begin{tabular}{@{}lccccccccccccccccccccc@{}}
\toprule
 & \multicolumn{4}{c}{TP} & \multicolumn{3}{c}{FP} & \multicolumn{7}{c}{TN} & \multicolumn{7}{c}{FN} \\ 
 \cmidrule(r){2-5}  \cmidrule(l){6-8} \cmidrule(l){9-15} \cmidrule(l){16-22}
 & $+$ & $\sim$ & $-$ & Total & $\sim$ & $-$ & Total & $++$ & $+\sim$ & $+-$ & $\sim$$\sim$ & $\sim-$ & $--$ & Total & $++$ & $+\sim$ & $+-$ & $\sim$$\sim$ & $\sim-$ & $--$ & Total \\ \midrule
ExDDI-IC & 0 & 2 & 1 & \textbf{3} & 6 & 3 & \textbf{9} & 2 & 0 & 1 & 0 & 0 & 1 & \textbf{4} & 0 & 0 & 0 & 0 & 0 & 4 & \textbf{4} \\
ExDDI-RV & 2 & 2 & 2 & \textbf{6} & 0 & 1 & \textbf{1} & 0 & 3 & 7 & 0 & 0 & 1 & \textbf{11} & 0 & 0 & 2 & 0 & 0 & 0 & \textbf{2} \\
ExDDI-S2S & 2 & 3 & 0 & \textbf{5} & 0 & 0 & \textbf{0} & 0 & 2 & 10 & 0 & 0 & 0 & \textbf{12} & 0 & 0 & 3 & 0 & 0 & 0 & \textbf{3} \\
ExDDI-MT & 2 & 3 & 0 & \textbf{5} & 0 & 0 & \textbf{0} & 2 & 3 & 7 & 0 & 0 & 0 & \textbf{12} & 0 & 0 & 3 & 0 & 0 & 0 & \textbf{3} \\
ExDDI-MTS & 2 & 3 & 1 & \textbf{6} & 0 & 0 & \textbf{0} & 2 & 3 & 7 & 0 & 0 & 0 & \textbf{12} & 0 & 0 & 2 & 0 & 0 & 0 & \textbf{2} \\ \bottomrule
\end{tabular}%
}
\caption{Results of human evaluation on the DDInter inductive test S2 set.}
\label{tab:human-s2}
\end{table*}

\begin{table*}[htb]
\resizebox{1.97\columnwidth}{!}{%
\begin{tabular}{@{}lccccccccccccccccccccc@{}}
\toprule
 & \multicolumn{4}{c}{TP} & \multicolumn{3}{c}{FP} & \multicolumn{7}{c}{TN} & \multicolumn{7}{c}{FN} \\ 
 \cmidrule(r){2-5}  \cmidrule(l){6-8} \cmidrule(l){9-15} \cmidrule(l){16-22}
 & $+$ & $\sim$ & $-$ & Total & $\sim$ & $-$ & Total & $++$ & $+\sim$ & $+-$ & $\sim$$\sim$ & $\sim-$ & $--$ & Total & $++$ & $+\sim$ & $+-$ & $\sim$$\sim$ & $\sim-$ & $--$ & Total \\ \midrule
ExDDI-IC & 0 & 0 & 2 & \textbf{2} & 7 & 1 & \textbf{8} & 1 & 0 & 0 & 0 & 0 & 2 & \textbf{3} & 0 & 1 & 1 & 1 & 2 & 2 & \textbf{7} \\
ExDDI-RV & 2 & 1 & 1 & \textbf{4} & 0 & 2 & \textbf{2} & 1 & 1 & 0 & 1 & 4 & 2 & \textbf{9} & 0 & 0 & 3 & 0 & 0 & 2 & \textbf{5} \\
ExDDI-S2S & 3 & 2 & 0 & \textbf{5} & 0 & 1 & \textbf{1} & 0 & 0 & 1 & 4 & 3 & 2 & \textbf{10} & 0 & 0 & 0 & 0 & 1 & 3 & \textbf{4} \\
ExDDI-MT & 1 & 1 & 1 & \textbf{3} & 0 & 3 & \textbf{3} & 0 & 0 & 0 & 3 & 2 & 3 & \textbf{8} & 0 & 0 & 1 & 0 & 3 & 2 & \textbf{6} \\
ExDDI-MTS & 1 & 1 & 3 & \textbf{5} & 0 & 3 & \textbf{3} & 0 & 0 & 0 & 3 & 2 & 3 & \textbf{8} & 0 & 0 & 1 & 0 & 2 & 1 & \textbf{4} \\ \bottomrule
\end{tabular}%
}
\caption{Results of human evaluation on the DDInter inductive test S1 set.}
\label{tab:human-s1}
\end{table*}

\onecolumn
\begin{xltabular}{\textwidth}{>{\centering\arraybackslash}m{0.1\textwidth}|p{0.85\textwidth}}
\toprule
\multicolumn{2}{c}{TP} \\ \midrule
$+$& \textbf{Drugs}: Perflutren; Nilotinib\newline \textbf{Gold Explanation}:  DRUG1 lipid microsphere suspension (Definity)has been reported to cause prolongation of the QT interval. Theoretically, coadministration with other agents that can prolong the QT interval may result in additive effects and increased risk of ventricular arrhythmias including torsade de pointes and sudden death. \newline \textbf{Model Explanation}:  DRUG2 can cause prolongation of the QT interval. Theoretically, coadministration with other agents that can prolong the QT interval may result in additive effects and increased risk of ventricular arrhythmias including torsade de pointes and sudden death. \newline \textbf{Model}: ExDDI-S2S (Inductive S1)\\\midrule

$\sim$& \textbf{Drugs}: Imatinib; Retapamulin\newline \textbf{Gold Explanation}:  Coadministration with inhibitors of CYP450 3A4 may increase the plasma concentrations of DRUG2, which is primarily metabolized by the isoenzyme. However, dosage adjustments are not necessary due to the low systemic exposure to DRUG2 following topical application. \newline \textbf{Model Explanation}:  When administered orally, DRUG2 may increase the plasma concentrations of drugs that are primarily metabolized by CYP450 3A4. Based on interaction with midazolam, a sensitive CYP450 3A4 substrate, DRUG2 may be a moderate CYP450 3A4 inhibitor. When oral midazolam was administered concomitantly with and at 2 or 4 hours after administration of DRUG2 tablets, mean midazolam peak plasma concentration (Cmax) and systemic exposure (AUC) increased by approximately 100\% and 200\%, respectively.
\newline \textbf{Model}: ExDDI-S2S (Inductive S1)\\\midrule

$-$& \textbf{Drugs}: Diazoxide; Ertugliflozin\newline \textbf{Gold Explanation}:  Sodium-glucose co-transporter 2 (SGLT-2) inhibitors may potentiate the hypotensive effects of diuretics and other antihypertensive agents or vasodilators. Inhibition of glucose and sodium co-transport produces mild diuresis and transient natriuresis, resulting in intravascular volume contraction. Volume depletion-related adverse reactions including hypotension, postural dizziness, orthostatic hypotension, syncope, and dehydration can occur after initiating treatment with SGLT-2 inhibitors, and the risk may be increased with concomitant use of other agents that can lower blood pressure. \newline \textbf{Model Explanation}:  The efficacy of insulin and other antidiabetic agents may be diminished by certain drugs, including atypical antipsychotics, corticosteroids, diuretics, estrogens, gonadotropin-releasing hormone agonists, human growth hormone, phenothiazines, progestins, protease inhibitors, sympathomimetic amines, thyroid hormones, L-asparaginase, alpelisib, copanlisib, danazol, diazoxide, isoniazid, megestrol, omacetaxine, phenytoin, tagraxofusp, temsirolimus, as well as pharmacologic dosages of nicotinic acid and adrenocorticotropic agents. These drugs may interfere with blood glucose control because they can cause hyperglycemia, glucose intolerance, new-onset diabetes mellitus, and/or exacerbation of preexisting diabetes. \newline \textbf{Model}: ExDDI-MT (Inductive S1)\\\midrule

\multicolumn{2}{c}{FP} \\ \midrule

$\sim$& \textbf{Drugs}: Daunorubicin; Zinc sulfate\newline \textbf{Gold Explanation}:  A very toxic anthracycline aminoglycoside antineoplastic isolated from Streptomyces peucetius and others, used in treatment of leukemia and other neoplasms. DRUG2 is a drug used to replenish low levels of zinc or prevent zinc deficiency, or to test for zinc deficiency. There were no known direct interactions reported between them. \newline \textbf{Model Explanation}:  DRUG1 is a medication used to treat various cancers and Kaposi's Sarcoma. DRUG2 is a zinc supplement used to prevent or treat zinc deficiency. There is a potential drug-drug interaction between these two molecules as zinc supplements may decrease the absorption of certain medications, potentially reducing the effectiveness of DRUG1. It is advisable to separate the administration of these two drugs by a few hours to minimize any potential interaction.
\newline \textbf{Model}: ExDDI-IC (Inductive S1)\\\midrule

$-$& \textbf{Drugs}: Artesunate; Dabrafenib\newline \textbf{Gold Explanation}:  DRUG1 is DRUG1 is an artemesinin derivative indicated for the initial treatment of severe malaria. DRUG2 is a kinase inhibitor used to treat patients with specific types of melanoma, non-small cell lung cancer, and thyroid cancer. There were no known direct interactions reported between them.
 \newline \textbf{Model Explanation}:  TCoadministration with inhibitors of CYP450 3A4 may increase the plasma concentrations of DRUG1 and lumefantrine, both of which are primarily metabolized by the isoenzyme. Coadministration with DRUG1 may decrease the plasma concentrations of CYP450 3A4 substrates such as protease inhibitors, macrolide antibiotics, and azole antifungal agents.
 \newline \textbf{Model}: ExDDI-MTS (Inductive S1)\\\midrule

\multicolumn{2}{c}{TN} \\ \midrule

$++$ & \textbf{Drugs}: Phenelzine; Regadenoson\newline \textbf{Gold Explanation}:  DRUG1 is a monoamine oxidase inhibitor used to treat atypical, nonendogenous, or neurotic depression. DRUG2 is a coronary vasodilator used in radionuclide myocardial perfusion imaging (MPI). There were no known direct interactions reported between them. \newline \textbf{Model Explanation}:  DRUG1 is a monoamine oxidase inhibitor used to treat major depressive disorder. DRUG2 is a coronary vasodilator used in radionuclide myocardial perfusion imaging (MPI). There were no known direct interactions reported between them.
\newline \textbf{Model}: ExDDI-MTS (Inductive S2)\\\midrule

$+\sim$ & \textbf{Drugs}: Nilotinib; Vitamin A\newline \textbf{Gold Explanation}:  DRUG1 is a kinase inhibitor used for the chronic phase treatment of Chronic Myeloid Leukemia (CML) that is Philadelphia chromosome positive and for the treatment of CML that is resistant to therapy containing imatinib. DRUG2 is a vitamin important for retinal function that is used clinically to correct DRUG2 deficiency. There were no known direct interactions reported between them. \newline \textbf{Model Explanation}:  DRUG1 is an antineoplastic kinase inhibitor used to treat chronic lymphocytic leukemia (CLL), relapsed follicular B-cell non-Hodgkin lymphoma (FL), and relapsed small lymphocytic lymphoma (SLL). DRUG2 is a vitamin important for retinal function that is used clinically to correct DRUG2 deficiency. There were no known direct interactions reported between them.
\newline \textbf{Model}: ExDDI-RV (Inductive S2)\\\midrule

$+-$ & \textbf{Drugs}: Naloxegol; Sucralfate\newline \textbf{Gold Explanation}:  DRUG1 is a peripherally-selective opioid antagonist used to treat opioid-induced constipation (OIC) in adult patients with chronic non-cancer pain. DRUG2 is a gastro-duodenal protective agent used in the treatment of gastric and duodenal ulcers and to prevent duodenal ulcer recurrence. There were no known direct interactions reported between them. \newline \textbf{Model Explanation}:  DRUG1 is a peripherally-selective opioid antagonist used to treat opioid-induced constipation (OIC) in adult patients with chronic non-cancer pain. DRUG2 is a sulfated pentosyl polysaccharide used to treat bladder pain and discomfort due to interstitial cystitis. There were no known direct interactions reported between them.
\newline \textbf{Model}: ExDDI-MTS (Inductive S2)\\\midrule

$\sim\sim$ & \textbf{Drugs}: Cefepime; Isoprenaline\newline \textbf{Gold Explanation}:  DRUG1 is a fourth-generation cephalosporin antibiotic used in the treatment of various bacterial infections caused by susceptible bacteria, such as pneumonia, urinary tract infections, and skin infections. DRUG2 is a catecholamine non-selective beta-adrenergic agonist typically used to treat bradycardia and heart block. There were no known direct interactions reported between them. \newline \textbf{Model Explanation}:  DRUG1 is a third-generation cephalosporin antibiotic used to treat bacterial infections of the respiratory and urinary tracts and of the skin and soft tissues. DRUG2 is a beta-2 adrenergic agonist used as a bronchodilator and to prevent premature labor. There were no known direct interactions reported between them.
\newline \textbf{Model}: ExDDI-S2S (Inductive S1)\\\midrule

$\sim-$ & \textbf{Drugs}: Daunorubicin; Zinc sulfate\newline \textbf{Gold Explanation}:  A very toxic anthracycline aminoglycoside antineoplastic isolated from Streptomyces peucetius and others, used in treatment of leukemia and other neoplasms. DRUG2 is a drug used to replenish low levels of zinc or prevent zinc deficiency, or to test for zinc deficiency. There were no known direct interactions reported between them. \newline \textbf{Model Explanation}:  DRUG1 is a medication used to treat various cancers and Kaposi's Sarcoma. DRUG2 is a medication used to treat convulsions during pregnancy, nephritis in children, magnesium deficiency, and tetany. There were no known direct interactions reported between them.
\newline \textbf{Model}: ExDDI-S2S (Inductive S1)\\\midrule

$--$ & \textbf{Drugs}: Phenelzine; Regadenoson\newline \textbf{Gold Explanation}:  DRUG1 is a monoamine oxidase inhibitor used to treat atypical, nonendogenous, or neurotic depression. DRUG2 is a coronary vasodilator used in radionuclide myocardial perfusion imaging (MPI). There were no known direct interactions reported between them.
 \newline \textbf{Model Explanation}:  DRUG1 is an ingredient of topical drugs for the relief of joint pain or muscle aches and pains. DRUG2 is a purine nucleoside analog and antineoplastic agent used for the treatment of with acute T-cell lymphoblastic leukemia and T-cell lymphoblastic lymphoma with inadequate clinical response to prior chemotherapeutic treatments. There were no known direct interactions reported between them.
\newline \textbf{Model}: ExDDI-RV (Inductive S2)\\\midrule

\multicolumn{2}{c}{FN} \\ \midrule

$+\sim$ & \textbf{Drugs}: Methyldopa; Iron \newline \textbf{Gold Explanation}:  INTERVAL: The oral bioavailability and pharmacologic effects of DRUG1 may be decreased during concurrent administration with DRUG2-containing products. The proposed mechanism is chelation of DRUG1 by the DRUG2 cation, forming an insoluble complex that is poorly absorbed from the gastrointestinal tract.
 \newline \textbf{Model Explanation}:  DRUG1 is a non-selective beta-adrenergic antagonist used in the treatment of hypertension, angina pectoris, and cardiac arrhythmias. DRUG2 is an essential mineral that is important for many bodily functions, including the transport of oxygen in the blood. There were no known direct interactions reported between them.
\newline \textbf{Model}: ExDDI-IC (Inductive S1)\\\midrule

$+-$ & \textbf{Drugs}: Lofexidine; Relugolix \newline \textbf{Gold Explanation}:  Long-term androgen deprivation therapy can prolong the QT interval. Theoretically, coadministration with other agents that can prolong the QT interval may result in additive effects and increased risk of ventricular arrhythmias including torsade de pointes and sudden death. In addition, the extent of drug-induced QT prolongation is dependent on the particular drug(s) involved and dosage(s) of the drug(s).
 \newline \textbf{Model Explanation}:  DRUG1 is an anthelmintic drug used to treat fascioliasis. DRUG2 is an oral GnRH receptor antagonist for androgen deprivation therapy in the treatment of advanced prostate cancer. There were no known direct interactions reported between them.
\newline \textbf{Model}: ExDDI-S2S (Inductive S2)\\\midrule

$\sim\sim$ & \textbf{Drugs}: Diazoxide; Ertugliflozin \newline \textbf{Gold Explanation}:  Sodium-glucose co-transporter 2 (SGLT-2) inhibitors may potentiate the hypotensive effects of diuretics and other antihypertensive agents or vasodilators. Inhibition of glucose and sodium co-transport produces mild diuresis and transient natriuresis, resulting in intravascular volume contraction. Volume depletion-related adverse reactions including hypotension, postural dizziness, orthostatic hypotension, syncope, and dehydration can occur after initiating treatment with SGLT-2 inhibitors, and the risk may be increased with concomitant use of other agents that can lower blood pressure.
 \newline \textbf{Model Explanation}:  There is no significant similarity in the molecular structures of Drug1 and Drug2 that would suggest a drug-drug interaction. Drug1 is a sulfonamide derivative with a benzothiazole ring, while Drug2 is a complex molecule containing a benzene ring and a sugar moiety. The functional groups and overall structures of the two drugs are quite different, indicating that they are unlikely to interact in a clinically significant manner.
\newline \textbf{Model}: ExDDI-IC (Inductive S1)\\\midrule

$\sim-$ & \textbf{Drugs}: Polythiazide; Selexipag \newline \textbf{Gold Explanation}:  Coadministration of medicines targeting the prostacyclin pathway (e.g., prostacyclin (PGI2), PGI2 analogs, or selective non-prostanoid prostacyclin IP receptor agonists) with diuretics, antihypertensive agents, or other vasodilators may potentiate the hypotensive effects of these agents. The mechanism involves the additive reduction in blood pressure due to the vasodilatory effects of PGI2, PGI2 analogs or selective non-prostanoid prostacyclin IP receptor agonists on the prostacyclin pathway.
 \newline \textbf{Model Explanation}:  DRUG1 is a sulfamyl diuretic used to treat edema in congestive heart failure, hepatic and renal disease, and nephrotic syndrome. DRUG2 is a class III antiarrhythmic drug used for the maintenance of normal sinus rhythm and cardioversion in cases of atrial fibrillation and atrial flutter. There were no known direct interactions reported between them.
\newline \textbf{Model}: ExDDI-MT (Inductive S1)\\\midrule

$--$ & \textbf{Drugs}: Siponimod; Upadacitinib \newline \textbf{Gold Explanation}:  Coadministration of DRUG1 with antineoplastic, immunosuppressive, or other immune-modulating therapies may increase the risk of unintended additive immunosuppressive effects.
 \newline \textbf{Model Explanation}:  There is no significant drug-drug interaction between Drug1 and Drug2 based on their molecular structures. The two drugs have distinct chemical structures and mechanisms of action, which do not suggest any potential interactions that would lead to adverse effects or altered pharmacokinetics.
\newline \textbf{Model}: ExDDI-IC (Inductive S2)\\

 \bottomrule
\caption{Examples of human evaluation. Drug names are provided in this table for readability, while models receive only SMILES representations as input.}
\label{tab:case-example}
\end{xltabular}

\twocolumn

\end{document}